\documentclass[11pt]{article}
\usepackage{coling2020}
\usepackage{times}
\usepackage{url}
\usepackage{latexsym}

\usepackage{graphicx}
\usepackage{hyperref}
\usepackage{float}
\usepackage{longtable}
\usepackage[normalem]{ulem}
\useunder{\uline}{\ul}{}
\usepackage{multirow}
\usepackage{hyperref}

\usepackage{comment}
\usepackage{float}

\usepackage{blindtext}

\colingfinalcopy % Uncomment this line for the final submission

\title{Definition Frames: Using Definitions for Hybrid Concept Representations}

\author{
Evangelia Spiliopoulou \qquad
Artidoro Pagnoni  \qquad 
Eduard Hovy\\
Language Technologies Institute\\
Carnegie Mellon University\\
\texttt{\{espiliop,apagnoni,hovy\}@cs.cmu.edu}}

\date{}

\begin{document}

\maketitle
\begin{abstract}
  Advances in word representations have shown tremendous improvements in downstream NLP tasks, but lack semantic interpretability. In this paper\footnote{Code available in  \href{https://github.com/spilioeve/Definition-Frames}{github.com/spilioeve/Definition-Frames}.}, we introduce Definition Frames (DF), a matrix distributed representation extracted from definitions, where each dimension is semantically interpretable. DF dimensions correspond to the Qualia structure relations \cite{boguraev1990lexical}: a set of relations that uniquely define a term. Our results show that DFs have competitive performance with other distributional semantic approaches on word similarity tasks.
\end{abstract}

\section{Introduction}

Ontologies have been widely used in lexical semantics to organize and represent knowledge. Carefully built by experts, they contain semantically meaningful information in the form of relations between concepts. However, being manually constructed, they struggle to assimilate new information.

Compared to ontologies, distributed representations are fully automated and can be fine-tuned for new tasks. Despite their exceptional performance, most distributional methods do not have an explicit semantic interpretation. The resulting representations encode a tremendous amount of information, but afford no way to interpret what this information is and how it relates to the concept. Thus, one cannot choose which type of information is useful for a specific task, unless one has a lot of data and resources to fine-tune. Although a few approaches have tried to bridge the gap between semantics and distributed representations \cite{faruqui2015retrofitting,mrkvsic2017semantic}, (1) they only encode information from ontologies, which are not extensible, and (2) the final representations are still not semantically meaningful.

 %However, most of those representations do not account for ambiguity  which we will discuss in more detail in the section \ref{2}.

Motivated by these problems, we introduce a novel hybrid representation called \textbf{Definition Frames} (DF), which encode semantic information extracted from definitions. DFs are matrix representations, where each row corresponds to a particular relation. The set of the relations used is based on the Qualia structure suggested in Boguraev and Postojovsky \shortcite{boguraev1990lexical}, and they are extracted automatically from definitions via a domain-adaptation approach. To the best of our knowledge, DF is the first hybrid representation, combining an explicit structure through semantically meaningful rows, while still being decomposed into distributional vectors. 
%Our contributions include: (1) a novel hybrid representation that is both semantically meaningful and can perform equally well in downstream tasks, (2) a computational approach to automatically extract the Qualia structure of any given entity and (3) an analysis of how each Qualia Structure might affect positively or negatively word similarity tasks. 

\section{Prior Work}
\label{2}
%While lexical semantics focuses on the classification and decomposition of a term with respect to its meaning, information extraction focuses on automatically extracting representations of the term from natural language text in the form of word embeddings or relations. 

Prior research on lexical semantics has established a set of relations that are sufficient to uniquely define a concept. Such work includes the Qualia structure \cite{boguraev1990lexical} and the generative lexicon theory \cite{pustejovsky1991generative}. Other related work includes ontological approaches \cite{baker1998berkeley,miller1995wordnet,Lenat:1995:CLI:219717.219745,speer2012representing} and more fine-grained definition-based frames like Semagrams \cite{moerdijk2008frames}.

%, which consists of four relation categories: formal, constitutive, telic and agentive. 

In distributional semantics, approaches including GloVe \cite{pennington2014glove}, word2vec \cite{mikolov2013distributed}, and fastText \cite{bojanowski2017enriching} obtain generic word embeddings by pre-training on large corpora. Recent work focused on context-sensitive embeddings like ELMo \cite{peters2018deep} and BERT \cite{devlin2018bert}, which achieve significant improvements in downstream NLP tasks.

%Dictionary definitions constitute an excellent source of human knowledge, as they contain essential relations about a concept. Although written in natural language, definitions follow a very specific structure. Most definitions of a concept contain the type to which it belongs (\textit{Genus}) and the properties that differentiate it from other concepts of the same class (\textit{Differentia}). In addition to their structure, definitions contain generic information that is sufficient to uniquely identify a concept, whereas most  natural language text (i.e. news articles, books, online forums) typically contain information about specific instances of a concept.  Those interesting properties of definitions motivate a series of work that uses them as sources to extract knowledge. 

%An excellent resource of human knowledge that contains essential relations about concepts is dictionary definitions. Definitions follow a specific structure and contain generic information about a concept, whereas most natural language text typically refers to specific instances. 
Earlier work on definitions extracted the type of a concept (\textit{Genus}) and the relations distinguishing it from other members of the same type (\textit{Differentia}) via syntax and string matching heuristics \cite{binot1993semantic,calzolari1984detecting,chodorow1985extracting}. Recent approaches directly encoded definitions to distributed representations. Tissier \shortcite{tissier2017dict2vec} obtained embeddings via a skip-gram model trained on definitions, while Bosc \shortcite{bosc2018auto} used an auto-encoder.
Other work includes definition generation \cite{noraset2017definition}, binary classification of sentences on whether they are definitional \cite{anke2018syntactically}, reverse dictionary look-up \cite{hill2016learning,Zock:2004:WLB:1610042.1610048}, and extraction of hypernymy relations from definitions using syntactic patterns \cite{boella2013extracting}.

%Other work focuses on enriching word embeddings with semantic knowledge from lexical resources. Examples include a process called Retrofitting, where they use belief-propagation to update embeddings on a relation graph \cite{faruqui2015retrofitting} and Counter-fitting \citeauthor{mrkvsic2017semantic, vulic2018specialising}, where they inject antonymy and synonymy constraints into embeddings.

%Another line of work focuses on enriching word embeddings with semantic knowledge from lexical resources, in a post-processing manner. \citeauthor{faruqui2015retrofitting} propose Retrofitting, a process where they use belief-propagation to update embeddings on a relation graph from a large ontology. Other work includes Counter-fitting \citeauthor{mrkvsic2017semantic,mrkvsic2016counter, vulic2018specialising}, where they inject antonymy and synonymy constraints into word embeddings. 

%\citeauthor{mrkvsic2017semantic} and \citeauthor{mrkvsic2016counter} inject antonymy and synonymy constraints into word embeddings, a process they call Counter-fitting. An interesting example of Counter-fitting is the LEAR framework, where they discuss the particular importance of the \textit{isA} relation in word embeddings \cite{vulic2018specialising}. 

\section{Approach}

Our framework consists of two parts: the Relation Retriever and the Definition Frame (DF) Encoder. The WordNet definition for any given term is used by the Relation Retriever model to extract the Qualia structure relations. The set of extracted terms pertaining to these relations form the Definition Frame. The DF Encoder encodes this output to a distributed matrix representation, which can be used in downstream NLP tasks. 

%\subsection{Qualia Structure}
\paragraph{Qualia Structure}
\label{3.1}

%While lexical semantics studies how word meaning can be decomposed, relation extraction focuses on detecting a set of important relations between terms. 

%scale=0.17
\begin{figure*}
   \centering
   \includegraphics[trim=20 400 0 300, width=\textwidth]{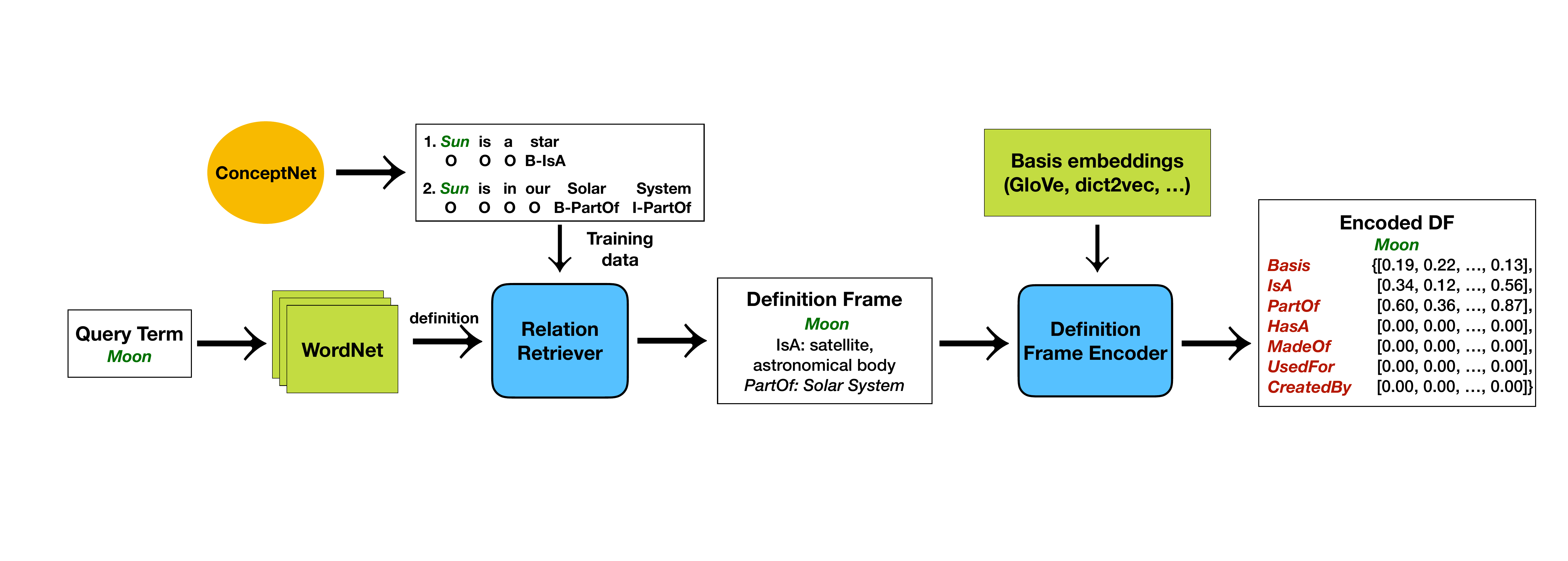}
   \caption{Architecture diagram.}
   \label{fig:Figure2}
\end{figure*} 

%Besides domain-specific relations, most Relation Extraction tasks \cite{gabor2018semeval,hendrickx2009semeval} contain relations describing the type (\textit{isA}), structure (\textit{madeOf, partOf, hasA}), function (\textit{usedFor}), or provenance (\textit{createdBy}) of a concept. However, these datasets This set of relation categories corresponds to the Qualia structure (formal, constitutive, telic, and origin), which is defined as the complete modes of explanation associated with an entity \cite{boguraev1990lexical,pustejovsky1991generative}. These relations usually suffice to uniquely define a concept and completely (more details in Appendix A). 

The Qualia structure (formal, constitutive, telic, and origin) is defined as the complete modes of explanation associated with an entity \cite{boguraev1990lexical,pustejovsky1991generative}. These relations suffice to uniquely and completely define a concept. 
In fact, several Relation Extraction tasks \cite{hendrickx2009semeval,gabor2018semeval} contain relations similar to Qualia describing the type (\textit{isA}), structure (\textit{madeOf, partOf, hasA}), function (\textit{usedFor}), or provenance (\textit{createdBy}) of a concept. 

\begin{table}[ht!]
% \small\addtolength{\tabcolsep}{-0.01pt}
\centering
\begin{tabular}{ |l | c | c| c| c|}
\hline
\textbf{Qualia} &\textbf{Relation} & \textbf{\# Wikipedia Def.}& \textbf{\# WordNet Def.} & \textbf{WordNet Overlap}\\
%&&\textbf{Sentences}\\
% && \textbf{Sent}& \textbf{Sent}\\
\hline
Formal & IsA & 235 & 146 & 59\% (87/146)\\
\hline
Constitutive / & PartOf & 82 & 57 & 2\% (1/57)\\
Structure & HasA & 39 & 33 & 6\% (2/33)\\
& MadeOf & 27 & 19 & 5\% (1/19)\\
\hline
Telic / &&&&\\
Function & UsedFor & 59 & 54 & 0\% (0/54)\\
\hline
Origin / &&&&\\
Provenance & CreatedBy & 26 &17 & 0\% (0/17)\\
\hline
% \textbf{Total Definitions} & & \textbf{300} & \textbf{150} &\\
% \hline
%\hline
%Total Entities & - & 263\\
%\hline
\end{tabular}
\caption{Annotated Relations for 300
Wikipedia and 150 WordNet definitions. \emph{WordNet Overlap} indicates the number of relations expressed in the definition that were present in the WordNet ontology.}
\label{annotated_rels}
\end{table}

To automatically extract the Qualia structure of a term, we use dictionary definitions, as they uniquely describe a term. We confirm the prevalence of those relations in definitions by annotating 300 Wikipedia and 150 WordNet definitions, chosen at random from nominal terms in WordNet (\autoref{annotated_rels}). We empirically find that WordNet definitions express more relations than the hypernymy (\textit{isA}) and meronymy (\textit{madeOf, partOf, hasA}) relations directly encoded in the WordNet ontology (usedFor and createdBy relations are not part of WordNet ontology). Furthermore, as shown in \autoref{annotated_rels}, we observe that meronymy relations are more prevalent in WordNet definitions compared to the ontology.

\paragraph{Training Data}
\label{3.2}
Because there are no definitions annotated with Qualia structure and Relation Extraction datasets \cite{hendrickx2009semeval,gabor2018semeval} are very domain specific without encoding general knowledge, we deploy a domain adaptation technique. We use ConceptNet to pre-train the Relation Retriever model (section \ref{3.3}) and then fine-tune it on and apply it to WordNet definitions. We fine-tune on a set of 150 manual annotations, since WordNet definitions tend to have more complex sentences than the ones in ConceptNet.  

ConceptNet \cite{speer2012representing} is a general purpose ontology that contains relations between pairs of concepts, accompanied by a small source-sentence. Figure \ref{fig:Figure2} shows that the Concept-query \textit{Sun} is linked to two sentences (\textit{Sun is a star} and \textit{Sun is in our solar system}) from ConceptNet with the corresponding relations \textit{isA} and \textit{partOf}. The training data for the Relation Retriever is composed of all ConceptNet source-sentences that contain one of the Qualia structure relations.

%In order to construct the training data, we extracted all ConceptNet relations that overlap with Qualia and the corresponding source-sentence. 

%However, most existing datasets on RE are particularly small or focus on a very narrow domain, which makes it hard to use them to obtain general relations. Given those constraints, we construct a large but simple dataset based on ConceptNet to pre-train the Relation Retriever model (more details in section \ref{3.3}).  

%We split our data into train (68,700 relations??correct it), dev and test (8,500 relations respectively).

%Wikipedia data is also processed in a similar way and used to construct the DF. One major difference compared to ConceptNet is that Wikipedia sentences are more complex, as they may contain relations of the Concept-query with multiple terms or even relations between terms other than the Concept-query.  

%\subsection{Extracting Definition Frames}
\label{3.3}
\paragraph{Extracting Definition Frames} 

The Relation Retriever uses the WordNet definition of a term to extract words that are related to that term via a Qualia-type relation. The set of extracted relations with their corresponding related words form the \textbf{Definition Frame} (DF). More specifically, we define a Definition Frame for a term $t$ as $F_t= \{r_1: S_1$, $r_2:S_2,$.., $r_k: S_k\}$,  where $r_i \in \{$ \textit{isA, usedFor, partOf, hasA, madeOf, createdBy} \} and $S_i$ is the set of words related to $t$ via the relation $r_i$. For example, to extract the DF for \textit{moon} (Figure \ref{fig:Figure2}), we use the WordNet definition of \textit{moon} as input. The Relation Retriever extracts the terms that are related to \textit{moon} via a Qualia-structure relation (i.e. \textit{satellite}, \textit{astronomical body} and \textit{solar system}). These terms with their corresponding relations constitute the Definition Frame $F_{moon}$. More examples of Definition Frames are shown in \autoref{tab:df}.

\begin{table}[ht!]
\small
\centering
%\small\addtolength{\tabcolsep}{-0.01pt}
\begin{tabular}{ | c | c |c | c| c| }
%\begin{tabular}{ | l | c |c |c| r| r| r| }
\hline
\textbf{Word 1} & \textbf{Definition Frame, word 1} & \textbf{Word 2} & \textbf{Definition Frame word 2} & \textbf{Relatedness}\\
\hline
shore & IsA: land, edge & sea  &IsA: body & 0.86\\
& PartOf: body, water && PartOf: ocean, salt, water &\\
& & &  CreatedBy: land &\\
\hline
wool & IsA: fabric & fabric & IsA: artifact &  0.86\\
& MadeOf: hair, sheep& &  MadeOf: weaving &\\
& & & HasA: fibers &\\
& & & CreatedBy: felting, knitting &\\
\hline
restaurant & IsA: building, people &dinner & IsA: main, meal& 0.86\\
& UsedFor: eat &&PartOf: day, evening, midday&\\
\hline
day& IsA: time & dusk &IsA: time & 0.76 \\
& UsedFor: earth, make, & &PartOf: day, following, sunset & \\
&complete, rotation&&&\\
\hline
dress &IsA: one-piece, garment & bride& IsA: woman & 0.76\\
&UsedFor: woman&&CreatedBy: married&\\
&HasA: skirt, bodice & &&\\
\hline
feather &IsA: light, horny, &hawk&IsA: diurnal, bird&0.82\\
& waterproof, structure&&HasA: short, rounded, &\\
&PartOf: external, covering&&wings&\\
\hline
orange& IsA: round, yellow, &fruit&IsA: ripened,& 0.82\\
&orange, fruit&&reproductive, body&\\
&PartOf: citrus, trees&&PartOf: seed, plant&\\
\hline
harbour & IsA: sheltered, port, ships& boat&IsA: small, vessel & 0.76\\
&UsedFor: discharge, cargo&&UsedFor: travel, water&\\
\hline
\end{tabular}
\caption{Extracted Definition Frames (before encoding) for pairs with high Relatedness score (MEN dataset). The Relatedness score, is the ground truth score, as noted in the original dataset. 
We observe that the two terms share characteristics of their Definition Frame, like being part of each other's frame or having common related terms.
}
\label{tab:df}
\end{table}

The Relation Retriever uses a BiLSTM model to extract the relations from each sentence. The task is formulated as a sequence tagging problem where we identify both the relation type and the related entities, and optimizes the cross-entropy loss. For model selection, we perform experiments with strong baseline architectures for RE tasks (BiLSTM, BERT-BiLSTM, BiLSTM-CNN). The Relation Retriever obtains F1 = 0.97 on ConceptNet test data (Appendix \ref{apx:relation_retriever}).

%\subsection{Definition Frame Encoder}
%\label{3.4}

The Definition Frame is encoded via the DF Encoder into a matrix where each row $w_i$ corresponds to one of the Qualia relations. The DF Encoder uses an embedding space ($Basis$) to construct each row vector $w_i$. Note that $Basis$ can be any distributional embedding model. Given a DF $F_t$, we define $w_i$ as the average of word embeddings from the set of related terms $S_i$ through relation $r_i$:
$$w_i = \frac{1}{|S_i|} \sum_{s \in S_i} Basis(s)$$
where $Basis(s)$ is the embedding for word $s$. We include an additional row for the $Basis$ vector of the term itself. This encoding of DF maintains a semantically meaningful structure as each row always corresponds to the same relation. If no terms are extracted for a relation, we use the zero vector of appropriate size. An example of the encoded DF$_{moon}$ is shown in Figure \ref{fig:Figure2}, where each dimension corresponds to a unique relation like \textit{isA} and \textit{partOf}.

\section{Experiments}

%\subsection{Word-Similarity Task}
\paragraph{Word-Similarity Task}
\label{4.1}

We perform experiments on benchmark word-similarity datasets provided by Faruqui \shortcite{faruqui-2014:SystemDemo}: SimLex999 \cite{hill2015simlex}, MC30 \cite{miller1991contextual},  RG65 \cite{rubenstein1965contextual}, WS353 \cite{finkelstein2002placing} and MEN \cite{bruni2012distributional}. 
Following Agirre \shortcite{agirre2009study}, we split them into word-similarity (WS-Sim, SimLex999, MC30, RG65) and word-relatedness (WS-Rel, MEN) datasets, as they evaluate different semantic affinities. We only consider nominal terms that exist in WordNet and report Spearman's correlation $\rho$. We perform experiments with three types of embeddings used as $Basis$: GloVe \cite{pennington2014glove}, dict2vec trained on Wikipedia \cite{tissier2017dict2vec}, and retrofit embeddings \cite{faruqui2015retrofitting} based on GloVe. Since the task comprises of pairs of words without any context, we do not compare against context-based representations.

%between the cosine similarity of the words representations and the normalized ground truth similarity score.

%Word-similarity tasks are particularly interesting, as words can be similar in different ways or facets. Although most of our data does not have an explicit type of similarity, we can divide them into two broad categories, as prior literature suggests: similarity and relatedness. For similarity datasets we use SimLex999 \cite{hill2015simlex} and MC-30 \cite{miller1991contextual}, while for relatedness we use MEN \cite{bruni2012distributional}, MTurk287 \cite{radinsky2011word}, MTurk771 \cite{halawi2012large} and RW-Stanford \cite{luong2013better}. Furthermore, we evaluate on WS-353 dataset \cite{finkelstein2002placing}  by dividing it into similarity and relatedness subsets (WS-SIM and WS-REL), as proposed by \citeauthor{agirre2009study}. 

%\subsection{Qualitative Analysis}

 %\subsection{Ablation Study}
 \paragraph{Ablation Study}
%(pruning dimensions that correspond to those relations)
We perform an ablation study by varying the set of relations used in DF. In this study, both $Basis$ and DF are encoded with dict2vec, as it achieves the best performance (Table \ref{tab:final}). The goal of this study is to measure how each extracted relation affects the performance of DF in word similarity tasks. The results (details in Appendix \ref{apx:ablation_study}) show that, for similarity tasks, pruning relations sometimes improves performance over both the original DF (with all relations) and the $Basis$ embeddings. However, we observe that DFs consistently have worse performance than $Basis$ in relatedness tasks, particularly in the MEN dataset. As we further discuss in detail in Section \ref{4.4}, although DFs capture relatedness, this is not reflected when using the cosine similarity metric directly, since it cannot compare information across different dimensions. For example, consider the pair (\textit{car}, \textit{wheel}). If we compare row-vectors of $DF_{wheel}$ and $DF_{car}$ for each relation separately, the representations are very different. Each Qualia structure relation defining \textit{car} and \textit{wheel} is different for the two terms. However, the Structure dimension of $DF_{car}$ would contain the information that \textit{wheel} is part (meronym) of \textit{car}, thus it should be compared to the $Basis$ dimension of $DF_{wheel}$.

\paragraph{Results}
\begin{table*}[h!]
%\begin{longtable}{| p{.20\textwidth} | p{.80\textwidth} |}
\centering
%\small\addtolength{\tabcolsep}{-0.01pt}
%\begin{tabular}{ | l | c |c | r| }
\resizebox{\columnwidth}{!}{%
\begin{tabular}{ |l || c | c || c  | c || c | c || c  | c || c | c || c  | c | }
\hline
\multicolumn{1}{|l||}{\textbf{Datasets}} &
 \multicolumn{4}{c ||}{\textbf{GloVe}} & \multicolumn{4}{c ||}{\textbf{Dict2vec}}& \multicolumn{4}{c |}{\textbf{Retrofit}}\\
% &Pr&Re& F1& P& R\\

\hline
 & Basis & Basis$^*$ & \textbf{DF} & \textbf{DF$^*$} &  Basis & Basis$^*$ & \textbf{DF} & \textbf{DF$^*$} & Basis & Basis$^*$ & \textbf{DF} & \textbf{DF$^*$}\\
\hline
Similarity CV&  0.39&  0.50 &  0.35 &  \textbf{0.53} &  0.53 &  0.52 &  0.45 &  \textbf{0.56} & 0.44 &  \textbf{0.59} &  0.35 &  0.56\\
%10-fold CV   &  &  &  & &  &  &  &  & & & &\\
\hline
Relatedness CV&  0.68 &  0.77 &  0.38 &  \textbf{0.80} &  0.71 &  0.76 &  0.61 &  \textbf{0.79} &  0.67 &  0.78 & 0.51 &  \textbf{0.80}\\
\hline
MEN-test&  0.70 &  0.79 &  0.56 &  \textbf{0.81} &  0.73 &  0.74 &  0.62 &  \textbf{0.79} &  0.71 &  0.79 &  0.53  &  \textbf{0.80}\\
%MEN$^*$ & 0.71 & 0.81 & 0.65 &  \textbf{0.83} & 0.73 & 0.80 & 0.65 & $ \textbf{0.82}^{*}$\\
\hline
\end{tabular}% 
}
\caption{Spearman's correlation for embeddings before and after the linear transform. All cross-validation (10-fold) experiments have p-value $p < 0.01$.}
\label{tab:final}
\end{table*}

To account for the cross-dimension problem described in the previous section, we design a slightly modified version of the previous experiments. 
We apply a linear transformation with the weights varying according to which type of word similarity (relatedness or similarity) we are measuring. 
%We use a simple technique of applying different weights in each dimension, according to the similarity-type being measured (similarity or relatedness).
%We apply a different linear transformation to the DF matrix viewed as a vector for each word similarity task -- relatedness and similarity.
This allows us to: (1) give more weight to more important relations and (2) compare the representations across different Qualia structure relations.

%In order to achieve this without substantially modifying the initial representations, we use a simple re-weighting of the dimensions. 
Using a linear transformation allows us to recover the initial DF representation from its transformed counterpart, which is important in order to maintain the semantic interpretability of DF (i.e. which words are related to $t$ and how). Thus, given $DF_t$ for a term $t$, we get $DF_t^{*} = W \times DF_t +b$, which we use in our experiments. The parameters $W$, $b$ are learnt separately for similarity and relatedness tasks, since different relations and cross-relation comparisons have varying importance for the two tasks. The training objective for the linear transformation is the minimization of the mean squared error between the cosine similarity of the transformed representations and the normalized ground truth similarity score. For fair comparison, we also apply a linear transformation to the baseline $Basis$ by learning parameters $W_\mathrm{basis}$, $b_\mathrm{basis}$ as described above for $DF$. In our experiments on similarity and relatedness datasets we use 10-Fold cross-validation and report the average performance, while on MEN we use the provided split into training and test data (it is the only dataset with a train/test split). 
%In order to validate our hypothesis about the effect of structure and whether the cosine similarity metric is an impediment for our representations, we design a slightly modified version of the previous experiments. For any dataset, instead of directly evaluating the encoded Definition Frame, we first apply a linear transformation on it. Thus, given the Definition Frames $DF_1$ and $DF_2$ for a pair of words $w_1, w_2$, we get

Our results show that Definition Frames achieve the best performance, compared to any of the baselines. In Table \ref{tab:final} we compare the performance of the Basis embeddings before and after the linear transformation ($Basis$ and $Basis^*$), with the Definition Frames ($DF$ and $DF^*$). $DF^*$ benefits much more of the dimension weighting and achieves better results compared to $Basis^*$, particularly with GloVe embeddings. Furthermore, we observe that Relatedness datasets (including MEN) gain the greatest advantage from the linear weighting. This lines up with our previous hypothesis, since the relatedness task requires more cross-relation comparisons ($DF_{car}$ vs $DF_{wheel}$).

\paragraph{Qualitative Analysis}
\label{4.4}
One of the distinguishing features of DFs is that they are semantically interpretable. Beyond determining whether two terms are related, we find that DFs can be used to infer \emph{how} they are related. We perform a qualitative analysis on 100 randomly selected terms from the MEN dataset that have high relatedness score (higher than 35 out of 50). The goal of this study is to assess whether we can use the explicit structure of DFs to predict the type of the relation between two terms. 
%To evaluate the quality of DFs and their ability to represent cross-dimension comparisons, we performed a qualitative analysis on terms from the MEN dataset. For this study, we used a subset of 100 randomly selected pairs that have high relatedness score. The goal is to assess whether we can use the explicit stucture of DFs to predict the type of the relation between two related terms. 

We conduct a Mechanical Turk study, where we present (1) the pair of related words, (2) their corresponding definitions and (3) a Qualia structure relation, in the form of question. We phrase the annotation task as a binary question such as ``\textit{Is an aquarium created by a fish?}''. We include all possible Qualia structure relations for each of the 100 pairs of related words. We ask three annotators to annotate each sample (1200 questions, each annotated three times, for a total of 3600 annotations).  

To identify the most probable relation between two terms $t_1$ and $t_2$ using the encoded DF, we conduct a set of row-to-row comparisons. We measure the cosine similarity of each row of $DF_{t_1}$ with $Basis(t_2)$ and vice-versa $DF_{t_2}$ with $Basis(t_1)$. The relation corresponding to the row with highest cosine similarity is taken to be the most probable relation.
%The output relation corresponds to the row with the highest cosine similarity score.
We test if the relation predicted by the DFs is correct according to humans. By taking the majority vote of the annotations, we find that 77\% of the extracted relations are considered valid by the workers. Furthermore, 54\% of the relations were considered accurate by all three annotators and the inter annotator percent agreement is 60\% over the 1200 relations (more details in Appendix \ref{apx:mturk}).

\section{Conclusion}

% Clayton proposal: We propose a hybrid representation whose dimensions, as grounded in lexical semantics, can also be employed as a distributed representation. 
We propose Definition Frames, a hybrid semantically interpretable representation that is grounded in both lexical semantics and distributed representations.
%by capturing different types of similarity (relatedness and similarity). 
By disentangling the Qualia structure relations, DFs can capture different types of similarity (relatedness and similarity) and achieve improved performance on word similarity tasks.
Finally, we demonstrate the explainability of Definition Frames via a human study showing that they provide valid insights on how terms are related. 
DFs are independent of the distributed representation used as basis. Future work could explore the use of contextual embeddings basis and the benefits of Definition Frames in downstream tasks.

%Finally, we demonstrate via a human study that DFs provide valid insights on how two terms are related, an important aspect of an explainable representation. 

%Future directions include improving the encoding of Definition Frames to a richer representation, exploring in depth how we can exploit the structure of Definition Frames to improve the representations (besides linear transformation) and using Definition Frames as a representation in downstream NLP tasks. 

\section*{Acknowledgements}

This research was partially supported by DARPA grant no HR001117S0017-World-Mod-FP-036 funded under the World Modelers program.
%Move that to Appendix??

\bibliographystyle{coling}
\bibliography{coling2020}

\newpage
\appendix
\section{Appendix}

\subsection{Relation Retriever performance}
\label{apx:relation_retriever}

In Table \ref{tab:rel_extr} we show the performance of the pre-trained Relation Retriever model on ConceptNet data, for all tested models. The performance is evaluated on a held-out test set. We observe that the performance is very high, which is our main motivation to fine-tune on the Qualia annotations of WordNet definitions.\\[15pt]

\begin{table}[ht!]
\centering
%\small\addtolength{\tabcolsep}{-0.01pt}
\begin{tabular}{ | c | c |c | c| }
%\begin{tabular}{ | l | c |c |c| r| r| r| }
\hline
\textbf{Model} & \textbf{Pr} & \textbf{Re} & \textbf{F1}\\
\hline
BiLSTM & 97.6 &97.7 &97.6 \\
\hline
BERT BiLSTM & 95.1& 95.0& 95.1\\
\hline
Stacked-BiLSTM &97.6&97.6&97.6 \\
\hline
BiLSTM-CNN&97.4&97.6 & 97.4 \\
\hline
\end{tabular}
\caption{Relation Retriever on ConceptNet data (held-out test set).}
\label{tab:rel_extr}
\end{table}

% \newpage
\subsection{Ablation Study}
\label{apx:ablation_study}
\begin{figure*}[h!]
  \centering
    \includegraphics[scale=0.75]{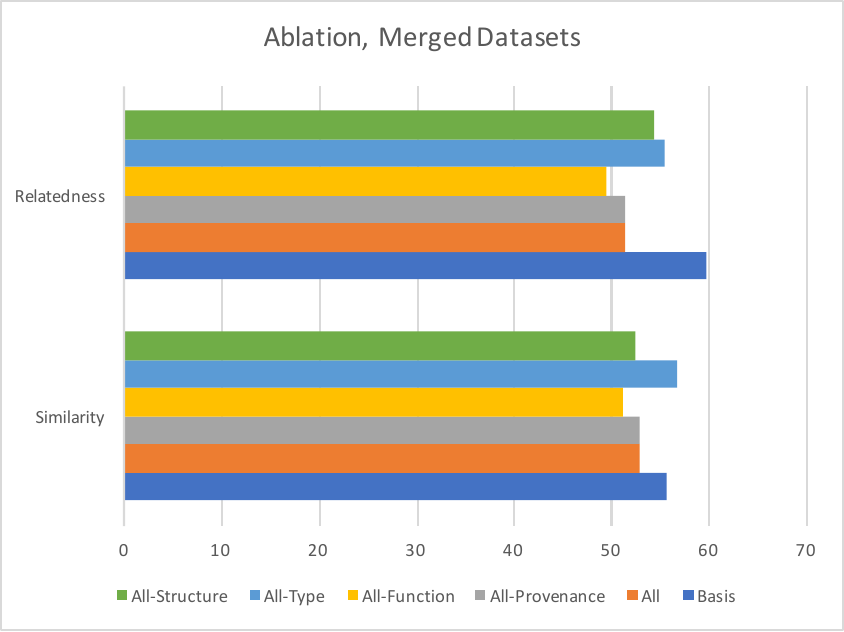}
    \caption{Ablation study for merged datasets.}
    \label{fig:Abl1}
\end{figure*}

\begin{figure*}[!ht]
  \centering
    \includegraphics[scale=0.75]{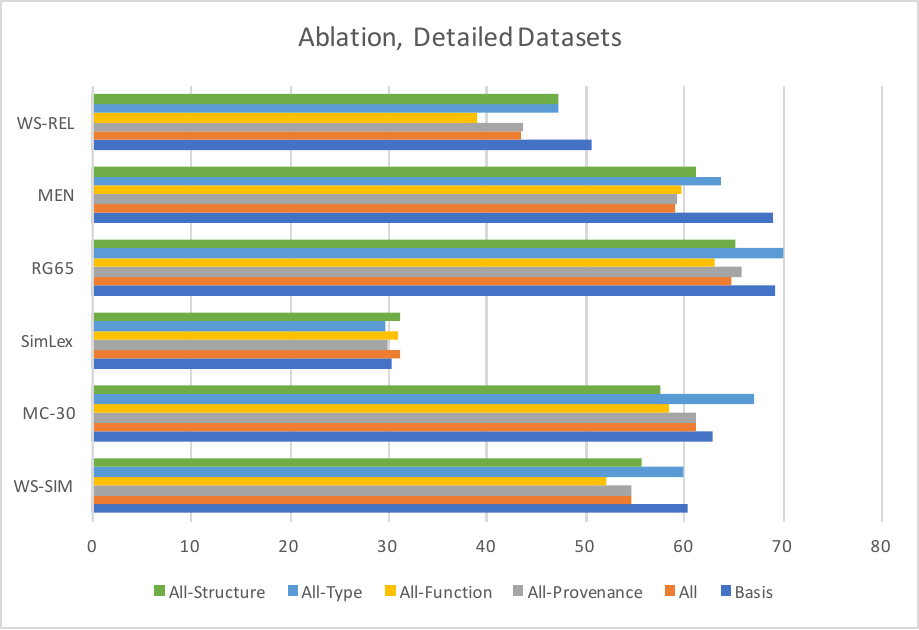}
    \caption{Ablation study for each dataset individually.}
    \label{fig:Abl2}
\end{figure*}

We compare the performance of $Basis$ embeddings with Definition Frames where one relation is pruned (\textit{All-$r$}, when relation $r$ is pruned). In Figure \ref{fig:Abl1} we show the ablation study when we merge the datasets into similarity and relatedness, while in Figure \ref{fig:Abl2}, we show the results of the study for each dataset separately.\\

\newpage
\subsection{MTurk Study Accuracy}
\label{apx:mturk}

In Table \ref{tab:mturk}, we show the accuracy per relation of the Definition Frames extracted relations, when all three MTurk participants agree.

\begin{table}[!ht]
\centering
%\small\addtolength{\tabcolsep}{-0.01pt}
\begin{tabular}{ | c | c | c| }
%\begin{tabular}{ | l | c |c |c| r| r| r| }
\hline
\textbf{Qualia} & \textbf{Relation} & \textbf{Agreement \%}\\
\hline
Formal & IsA & 0.43\\
\hline
Constitutive / & PartOf, & 0.79\\
Structure & HasA, & \\
& MadeOf & \\
\hline
Telic / &&\\
Function & UsedFor & 0.50\\
\hline
Origin / &&\\
Provenance & CreatedBy & 0.25\\
\hline
\end{tabular}
\caption{Accuracy per relation.}
\label{tab:mturk}
\end{table}

\end{document}